\documentclass[letterpaper]{article} 
\usepackage{aaai24}  
\usepackage{times}  
\usepackage{helvet}  
\usepackage{courier}  
\usepackage[hyphens]{url}  
\usepackage{graphicx} 
\usepackage{natbib}  
\usepackage{caption} 
\usepackage{lipsum}
\usepackage{algorithm}
\usepackage{algorithmic}
\usepackage{amsmath,amssymb}
\usepackage{bm}
\usepackage{booktabs}
\newcommand{\Lagr}{\mathcal{L}}
\newcommand{\mat}[1]{\bm{#1}}

\usepackage{multirow}
\usepackage{float}
\usepackage{color, colortbl}
\usepackage[cmyk]{xcolor}

\usepackage{cuted}

\usepackage{newfloat}
\usepackage{listings}
\DeclareCaptionStyle{ruled}{labelfont=normalfont,labelsep=colon,strut=off} 
\floatstyle{ruled}
\newfloat{listing}{tb}{lst}{}
\floatname{listing}{Listing}
\title{Learning Spatially Collaged Fourier Bases for Implicit Neural Representation}
\author{
    Jason Chun Lok Li\equalcontrib,
    Chang Liu\equalcontrib,
    Binxiao Huang,
    Ngai Wong
}
\affiliations{
   Department of Electrical and Electronic Engineering, The University of Hong Kong \\
   \{jasonlcl, lcon7, huangbx7\}@connect.hku.hk, nwong@eee.hku.hk
}

\usepackage{bibentry}

\begin{document}

\maketitle

\begin{abstract}
Existing approaches to Implicit Neural Representation (INR) can be interpreted as a global scene representation via a linear combination of Fourier bases of different frequencies. However, such universal basis functions can limit the representation capability in local regions where a specific component is unnecessary, resulting in unpleasant artifacts. To this end, we introduce a learnable spatial mask that effectively dispatches distinct Fourier bases into respective regions. This translates into collaging Fourier patches, thus enabling an accurate representation of complex signals. Comprehensive experiments demonstrate the superior reconstruction quality of the proposed approach over existing baselines across various INR tasks, including image fitting, video representation, and 3D shape representation. Our method outperforms all other baselines, improving the image fitting PSNR by over 3dB and 3D reconstruction to 98.81 IoU and 0.0011 Chamfer Distance.
\end{abstract}

\section{Introduction}
\label{sec:introduction}
Implicit Neural Representation (INR) has attracted increased attention lately as an effective framework for encoding complex signals in diverse fields, encompassing image fitting, 3D shape representation, and novel view synthesis~\cite{2020siren, saragadam2023wire, mildenhall2021nerf}. Using INR, discontinuous signals such as images, videos, and 3D point clouds can be represented by a lightweight neural network, which outputs corresponding data values given the query coordinates. Hence, the neural network transforms the original discontinuous data into a continuous latent mapping function, which sheds new light on several downstream tasks, for example, data compression, 3D shape reconstruction, and super-resolution~\cite{2020siren, saragadam2023wire}.

\begin{figure}[ht]
    \centering
    \includegraphics[width=1\columnwidth]{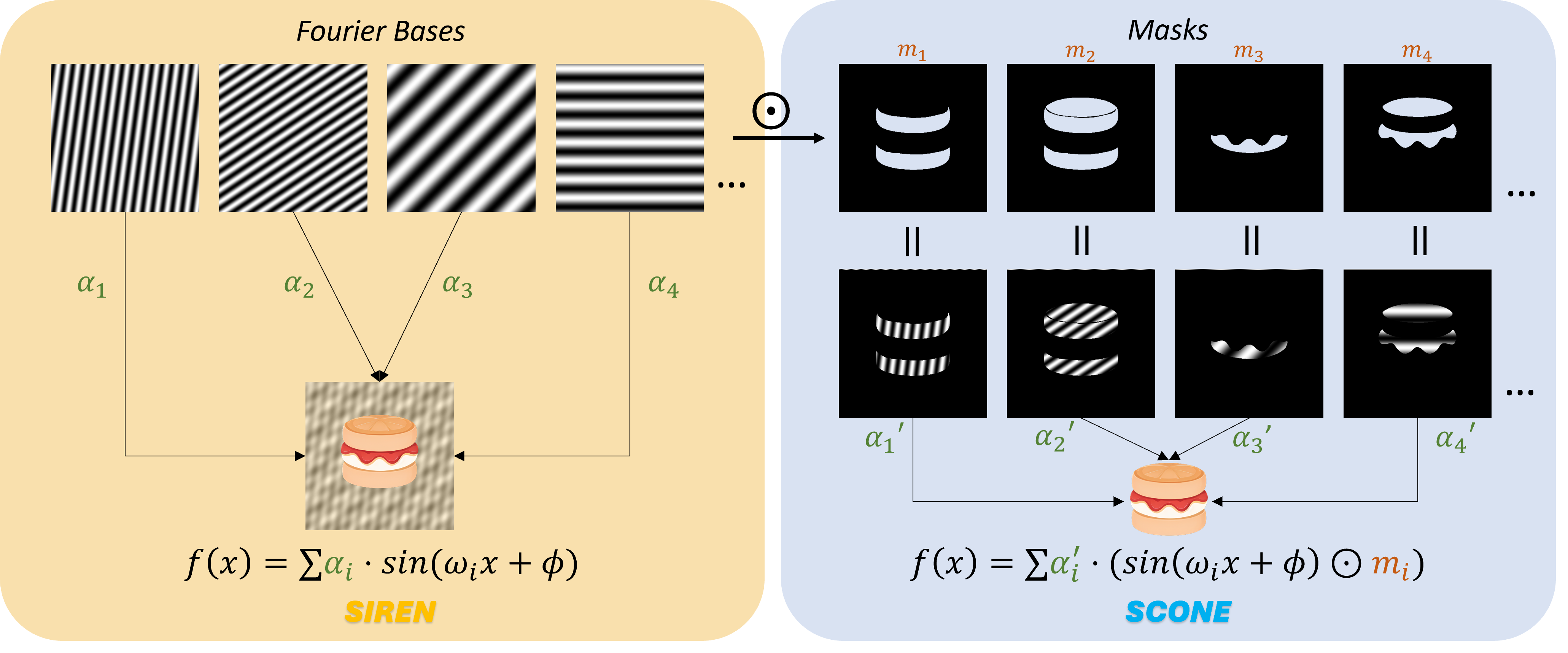}  
    \caption{\textbf{(Left)} traditional INR such as SIREN can be viewed as a linear combination of Fourier features of different frequencies, where the globally added Fourier features are universally applied to the whole frame even for some regions where these frequencies are unnecessary. \textbf{(Right)} The proposed SCONE utilizes spatial masks for each Fourier feature and collages the frequency patches to form a more precise reconstruction. }
    \label{fig:idea}
\end{figure}

Albeit the success of INR, conventional methods usually model the target signal through a linear combination of universal basis functions spanning various frequencies~\cite{2020siren, tancik2020fourier, fathony2020multiplicative}, which are usually waveform Fourier features. Consequently, these methods inevitably introduce the same strength of each basis feature to the whole space, even when the component is \textit{not} necessary for certain regions. Even though such unnecessary components can be counteracted by combining a large number of different frequency bases, this strains the network and limits its representability. As also pointed out by~\cite{dou2023multiplicative}, the global combination manner results in larger model sizes and/or sub-optimal performance, especially for local regions where existing methods can introduce undesirable artifacts and fuzzy details.

In view of such limitation, this paper presents a simple yet effective architecture, dubbed \textbf{S}patially \textbf{C}ollaged c\textbf{O}ordinate \textbf{NE}tworks (SCONE). In contrast to existing approaches employing a combination of global Fourier features, SCONE incorporates learnable spatial masks to allocate specific Fourier bases to their corresponding regions. As illustrated in Fig.~\ref{fig:idea}, SCONE learns a spatial mask for each Fourier feature to avoid adding the wave component in regions where it is unnecessary. Consequently, SCONE distinguishes between the distinct tasks of learning geometry-dependent masks and Fourier features, permitting the Fourier features to be spatially and adaptively decoupled. By adopting a different paradigm of spatial collaging instead of global overlaying, the proposed technique facilitates a more precise and granular representation of intricate signals through the efficient assembly of mask-cropped Fourier patches. Additionally, it has been observed that SCONE converges faster than baseline methods, suggesting the gradual assembly of Fourier patches can provide better learning dynamics. In short, SCONE effectively overcomes the limitations of existing global basis functions and offers superior reconstruction quality across a wide spectrum of tasks.

To demonstrate the efficacy of SCONE, we extensively evaluate SCONE on various INR tasks, including image fitting, video representation, and 3D shape representation, and compare its performance against existing baselines. The experimental results \textit{consistently} show the superiority of SCONE in terms of reconstruction quality, overtaking its competing counterparts. For image fitting, SCONE outperforms all  baselines by a significant margin over 3dB PSNR. It also achieves a new state-of-the-art (SOTA) performance of 98.81\% IoU and 0.0011 Chamfer Distance on 3D shape representation. Our contributions are threefold:
\begin{itemize}
    \item We identify the limitation in existing methods of globally combining Fourier bases and propose SCONE to decouple the bases spatially.
    \item SCONE is the first-ever coordinate network that explicitly collages patches of different frequencies.
    \item Comprehensive experiments are conducted to evaluate SCONE on various tasks. Our method consistently sets new SOTA on all tasks. 
\end{itemize}

\section{Related Work}
\label{sec:background}
Implicit Neural Representations (INRs), also known as coordinate networks, can encode continuous signals in a compact form using constant memory, depending only on model capacity, regardless of the resolution of their discrete representation (e.g., pixels or voxel grids). They take low-dimensional coordinates as inputs and are trained to learn the input-output mapping in an end-to-end fashion via gradient-based optimization. INRs have been applied to represent a wide variety of signals, including, but not limited to, images~\cite{saragadam2023wire,lindell2022bacon, shekarforoush2022residual}, occupancy networks~\cite{chen2019learning, mescheder2019occupancy}, signed distance functions~\cite{sitzmann2019scene, park2019deepsdf}, as well as solving differential equations~\cite{2020siren, fathony2020multiplicative}.
 
INRs first gained attention in the field of novel view synthesis~\cite{mildenhall2021nerf}, which uses a multilayer perceptron (MLP) to represent a 5D continuous radiance field (viz. neural radiance field), which maps radiance directions and spatial coordinates to view-dependent color and density. The model is trained using 2D images from various views, and a differential volume rendering equation is employed to render the MLP output for novel views. Positional encoding is utilized~\cite{vaswani2017attention} to represent high-frequency scene content effectively.

Following that, Ref.~\cite{tancik2020fourier} systematically shows that the positional encoding adopted is one type of Fourier-feature mapping that enables a coordinate-based MLP to overcome its inductive spectral bias of failing in learning high-frequency content. With the use of Neural Tangent Kernel (NTK)~\cite{jacot2018ntk} theory, the authors show that Fourier feature mapping can successfully transform the effective NTK of an MLP into a stationary kernel with a tunable bandwidth, thus improving its capability of fitting natural signals such as images. Apart from positional encoding, Gaussian Random Fourier Features (RFFs) are also proposed, which leads to improved reconstruction quality. 

\begin{figure*}[th]
    \centering
    \includegraphics[width=2\columnwidth]{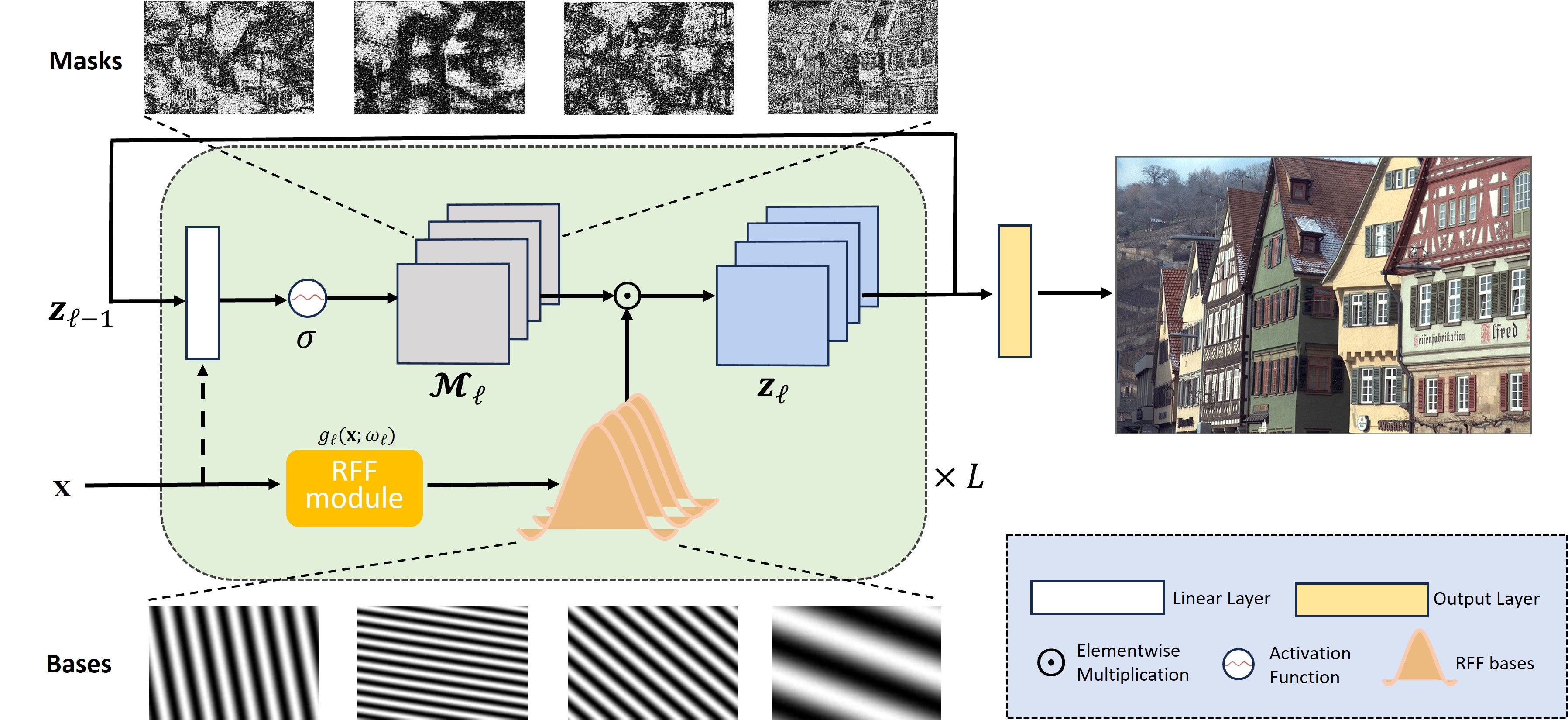}  
    \caption{An overview of SCONE architecture. The dashed arrow denotes that the initial hidden features $z_{0}$ are from the coordinates $x$. $\sigma$ is the activation function with output range $[0, 1]$. Independent Fourier bases in each level are masked to form Fourier patches that are then spatially collaged by the next linear layer. This enables the assignment of global bases to their proper spatial positions, which are then reused in more than one region across the signal.}
    \label{fig:architecture}
\end{figure*}

However, even when the coordinates are first transformed by positional encoding via Fourier features, the use of ReLU MLPs, whose second derivative is zero, limits the model's ability to represent the derivatives of a target signal. This constraint has motivated the development of sinusoidal representation networks (SIREN)~\cite{2020siren}. Specifically, SIREN replaces the ReLU activation with periodic sinusoidal nonlinearities, which can accurately represent natural signals and their derivatives. In addition to sine activations, Wavelet Implicit Neural Representation (WIRE)~\cite{saragadam2023wire} attempts to utilize a continuous complex Gabor wavelet activation function that is optimally concentrated in space-frequency. This provides appropriate inductive biases for representing images.

On the other hand, instead of stacking layers in a composite manner, multiplicative filter networks (MFNs)~\cite{fathony2020multiplicative}, a family of INRs with better interpretability, repeatedly apply linear functions of sinusoidal or Gabor filters to the input, which are then element-wise multiplied together. The entire expression can be rewritten as a linear combination of an exponential number of sinusoidal or Gabor wavelet bases. Two works have been developed to extend the theoretical understanding and practicality of MFNs. BACON~\cite{lindell2022bacon} is a variant of MFNs that has band-limited properties, resulting in an analytical Fourier spectrum. Whereas RMFN~\cite{shekarforoush2022residual} introduces residual connections and a novel initialization scheme, allowing more control of the frequency spectrum at each stage of optimization and enabling coarse-to-fine estimation.

By taking a step back, a recent work~\cite{yuce2022structured} attempts to provide an overarching explanation for the underlying theory behind INRs, including SIREN and MFNs. Results from harmonic analysis and NTK theory are utilized to demonstrate that most INR families are analogous to structured signal dictionaries. This structure enables INRs to express signals with an exponentially increasing frequency support using a number of parameters that grow linearly with depth.

Two approaches closely related to our work are GaborNet instantiation of MFN and WIRE. They enhance the locality of INRs by employing spatially compact wavelet filters to improve their ability to represent local regions. But rather differently, our proposed SCONE accurately represents local structures \textit{without} relying on the computationally intensive wavelet transform (which may involve complex number computations). We achieve this by collaging Fourier patches with the help of positional-dependent masks. A multi-channeled mask is applied to Fourier bases of varying frequencies in each stage, resulting in Fourier patches. These patches are then mixed using a linear layer. The resulting output is subsequently used as an input to generate the mask for the next stage. The whole process can be interpreted as making a collage by assembling different textures, as illustrated in Fig.~\ref{fig:idea}. Unlike Gabor filters or wavelet activations, where each wavelet waveform can only appear in specific spatial locations, the adaptive mask in SCONE can effectively dispatch global bases to multiple locations in the signal.

\section{Preliminary}
\label{sec:method_inr}
An INR or a coordinate network is a method that uses a neural network to encode continuous signals. The network is trained to learn the complex relationships between the coordinates and their corresponding features, in order to accurately encode the signal continuously. For example, an image can be described by a function that maps each pixel's coordinates $\mathbf{x} = (x, y)$ in $\mathbb{R}^2$ to its corresponding RGB value $\mathbf{y} = (r, g, b)$ in $\mathbb{R}^3$. With a single image datapoint as a collection of coordinate and feature pairs $\mathbf{d} = \{(\mathbf{x_i}, \mathbf{y_i})\}_{i=1}^n$, the INR fits a neural network $f_\theta : \mathbb{R}^m \rightarrow \mathbb{R}^c$ with parameters $\theta$ to the datapoint by minimizing the loss function:

\begin{equation}
\Lagr(\theta, \mathbf{d}) =  \sum^{n}_{i=1}\Vert f_\theta(\mathbf{x_i}) - \mathbf{y_i} \Vert_2
\end{equation}%
The conventional parameterization of an $L$-layer INR exhibit a general form:
\begin{equation}
    \begin{aligned}
         \mathbf{z}_{0} &= \gamma(\mathbf{x}) \\
         \mathbf{z}_{\ell} &= \delta(\mat{W}_{\ell} \mathbf{z}_{\ell-1} + \mathbf{b}_{\ell}) \ \ , \ \ \ell = 1, ... , L-1 \\
         \mathbf{z}_{L} &= \mat{W}_{L}\mathbf{z}_{L - 1} +\mathbf{b}_{L}
    \end{aligned}%
\end{equation}%
where $\mathbf{z}_{\ell}\in\mathbb{R}^{D_{\ell}}$ represents the post-activation latent feature, $\mat{W}_{\ell}\in\mathbb{R}^{D_{\ell}\times D_{\ell-1}}$ represents the weight matrix, and $\mathbf{b}_{\ell}\in \mathbb{R}^{D_{\ell}}$ represents the bias vector. $\gamma(\mathbf{x})$ denotes the input mapping, also known as positional encoding, and $\delta(\cdot)$ represents the nonlinear activation function. 

In particular, both Fourier feature networks (FFNs)~\cite{tancik2020fourier} and SIREN~\cite{2020siren} models can be expressed in such unified form. FFNs utilize Fourier feature mappings to lift the input coordinates into frequency space, followed by MLPs with ReLU nonlinearity. According to~\cite{tancik2020fourier}, using Gaussian RFFs $\gamma(\mathbf{x}) = [\cos(2\pi\mathbf{Bx}), \sin(2\pi\mathbf{Bx})]^T$, where $\mathbf{B} \in \mathbb{R}^{m \times d}$ is a random matrix sampled from $\mathcal{N}(0, \sigma^2)$, yields the best results. On the other hand, SIREN can also be interpreted as an MLP with sine nonlinearity and the first layer being a feature mapping with $\gamma(\mathbf{x}) = \sin{(\omega_0 (\mat{W}_{0} \mathbf{x} + \mathbf{b}_{0}))}$. Both $\sigma^2$ in FFNs and $\omega_0$ in SIREN are global hyperparameters controlling the signal frequency of the input mapping. 

\section{Spatially Collaged Coordinate Networks}
\label{subsec:method_scone}
This section elucidates SCONE. As described earlier, SCONE can be interpreted as a spatial-aware combination of Fourier bases with different frequencies. As shown in Fig.~\ref{fig:architecture}, we iteratively generate spatial masks from previous-stage latent features, which will be applied for collaging Fourier Features with different frequencies. Finally, the SCONE output is derived as a linear combination of the collaged RFFs.

\subsubsection{Random Fourier Features}
\label{subsubsec:method_rff}
Given the input coordinates $\mathbf{x}$, SCONE first normalizes the inputs into a range of $[-1, 1]^2$ and transforms them into position encoding using RFFs:
\begin{equation}
    g_{\ell}(\mathbf{x};\omega_{\ell}) = \sin(\omega_{\ell}(\mathbf{B_{\ell}x}))
\end{equation}

\noindent where $\mathbf{B_{\ell}}\sim U(-\sqrt{1/2}, \sqrt{1/2})$ is a random matrix. This results in $L$ independent Fourier bases, one for each layer controlled by the hyperparameter $\omega_{\ell}$ as depicted in the bottom part of Fig.~\ref{fig:architecture}.

\begin{figure*}[th]
    \centering
    \includegraphics[width=2\columnwidth]{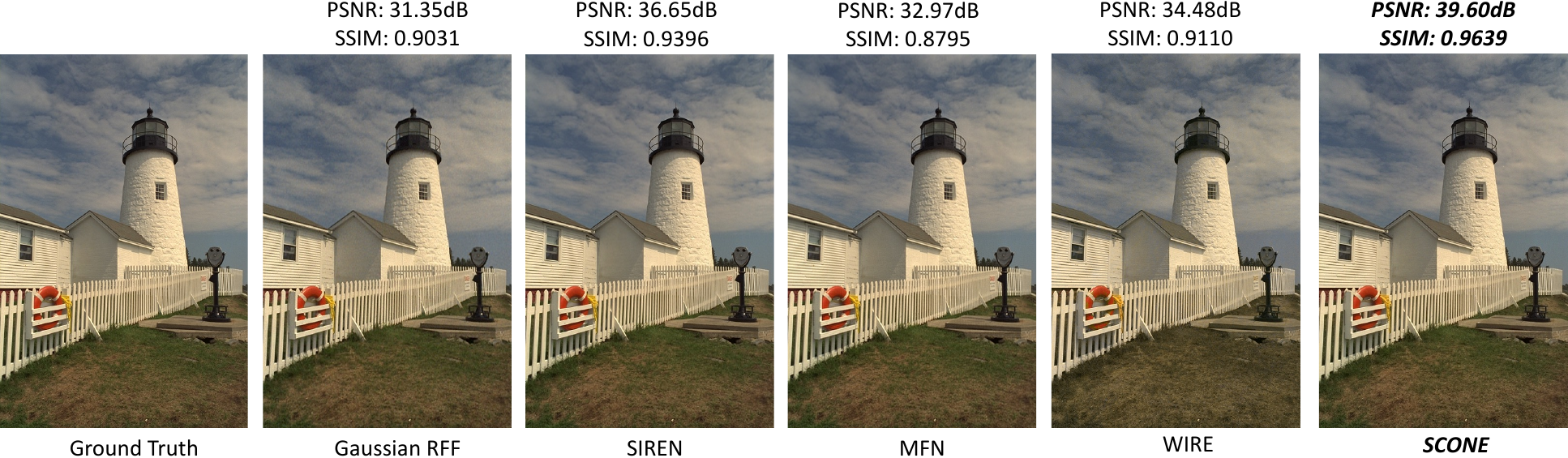}  
    \caption{The result on image representation task. Kodak19 is selected for comparison.}
    \label{fig:2d_results}
\end{figure*}

\begin{figure}[ht]
    \centering
    \includegraphics[width=0.9\columnwidth]{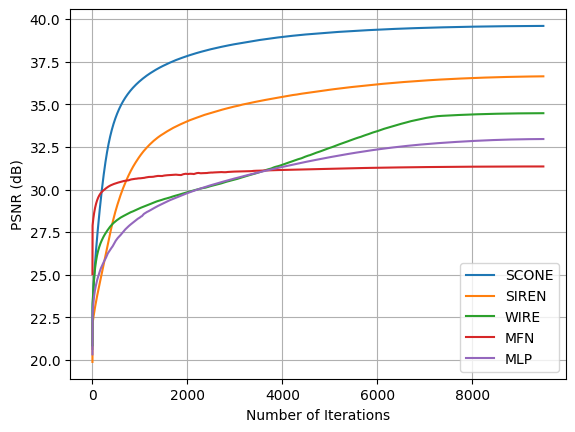}  
    \caption{PSNR vs. Training Iterations: The plot depicts the PSNR against the training iterations for the image fitting task (Kodak09). With the aid of geometry-dependent spatial masks, SCONE converges at a much faster rate and achieves the highest reconstruction accuracy.}
    \label{fig:psnr_vs_iters}
\end{figure}

\subsubsection{Spatial Mask}
\label{subsubsec:method_mask}

During the mask generation process, the input coordinates first pass through a linear mapping with a scaling factor as described as $\mathbf{z}_{0} = \omega_0 (\mat{W}_{0}\mathbf{x})$. Similar to SIREN, $\omega_0$ can be seen as a global parameter to control the frequency of the mask-generating process.

Iteratively, SCONE generates spatial masks for collaging RFF bases, thereby enabling various RFF combinations at different regions of the image. Specifically, given last layer latent feature $\mathbf{z}_{\ell-1}$, the mask generating process takes place in following form:
\begin{equation}
    \begin{aligned}
         \mathcal{M}_{\ell} &=\sigma(\mat{W}_{\ell-1}\mathbf{z}_{\ell-1} + \mathbf{b}_{\ell-1})) \\
         \mathbf{z}_{\ell} &=  \mathcal{M}_{\ell} \circ g_{\ell}(\mathbf{x};\omega_{\ell})    \ \ , \ \ \ell = 1, ... , L-1 \\
    \end{aligned}%
\end{equation}%


\noindent where $\sigma$ is an activation function and $\circ$ denotes element-wise multiplication and the weight matrix $\mat{W}_{\ell}\in\mathbb{R}^{D_{\ell}\times D_{\ell-1}}$ is initialized using the same initialization scheme as SIREN. 


By applying a corresponding spatial mask on each RFF, the frequency will only be added to certain region of the image instead of globally as in SIREN~\cite{2020siren}. Moreover, different from existing methods such as WIRE~\cite{saragadam2023wire}, instead of manipulating the activation function to shrink the RFFs into a closed region, our method allows one RFF to be used at multiple regions with irregular or complex shapes. Without extra supervision signal, SCONE learns to generate spatial masks aligning with the target image distribution in a highly interpretable manner. We visualize the generated masks in the upper part of Fig.~\ref{fig:architecture}. We also provide a visualization of hidden layers' activations in SCONE, as well as those of other models in the later Section for comparison.

In practice, the proposed model progressively learns simple to complex spatial masks. Taking advantage of the increasingly elaborate spatial masks, albeit using the same number of RFFs, it is observed that outputs from each iteration appear to contain recognizable shapes in deep layers. In the end, a fully connected layer generates the final output image as a linear combination of the last set of collaged Fourier bases.

\section{Experiments}
In this section, we evaluate our SCONE for various tasks of image fitting, video representation, and 3D shape representation. Our method is proven to surpass all the other baseline methods by a large margin. We also visualize the layer activations for interpretability.

\label{sec:experiments}
\subsection{Implementation Details}
All models are trained for 10k iterations on Nvidia RTX 3090 GPUs, each equipped with a 24GB memory buffer. The training process utilizes the Adam optimizer~\cite{kingma2014adam}, with the parameters $\beta_1 = 0.9$ and $\beta_2 = 0.999$, and employs the Mean Squared Error (MSE) loss without weight decay. Additionally, a cosine learning rate scheduler is applied, with a minimum learning rate of $1e-6$. We evaluate SCONE against several baselines, including ReLU MLP with Gaussian RFF encoding~\cite{tancik2020fourier}, SIREN~\cite{2020siren}, GaborNet instantiation of MFN~\cite{fathony2020multiplicative}, and WIRE~\cite{saragadam2023wire}. The implementation of all codes is carried out using the PyTorch~\cite{pytorch} framework, and the baselines are based on official codes released by authors of the respective models. For the below experiments, we choose $\sin^2$ as the activation function $\sigma$ due to its desirable output range of $[0,1]$ for generating soft masks and periodic property. In a later section, we conduct a thorough analysis of the impact of different activation functions, which demonstrates that SCONE is robust to the choice of activation function, not limited to $\sin^2$. The excellent performance of SCONE can be attributed to the design of collaging Fourier features. Please refer to the ablation study for more details.


\subsection{Image Fitting}
\label{subsec:exp_image}
The image representation task is performed on selected images from the Kodak dataset~\cite{kodak}, in which each image has a resolution of $512\times768$ or $768\times512$. Two widely adopted metrics, namely Peak Signal-to-Noise Ratio (PSNR) and structural similarity index (SSIM)~\cite{wang2004image}, are used as evaluation metrics. To align the model sizes, all models have 256 hidden units in each layer, except for SCONE and WIRE, which has 250 and 360 hidden units, respectively. We find that naively increasing the number of layers in WIRE leads to inferior performance. To fix the number of parameters to around 200k, both SIREN and SCONE have 5 layers, while the others have 4. We conduct hyperparameter sweeps for each model to select the best hyperparameters and initial learning rate. Fig.~\ref{fig:2d_results} demonstrates that SCONE outperforms all baselines by a significant margin ($>3$dB). The ReLU MLP with Gaussian RFF exhibits high-frequency noise on relatively flat surfaces. MFN displays severe ringing artifacts around the edges. WIRE struggles with learning color, resulting in poor saturation (as seen in the grass). Both SIREN and SCONE achieve good reconstruction, with SCONE producing sharper and finer textures (as seen in the house next to the tower). We also compare the convergence speed of SCONE with other baselines in Fig.~\ref{fig:psnr_vs_iters}. It can be seen that MFN has a fast initial convergence speed but quickly becomes flat, failing to learn high-quality representations. In our setting with 200k parameters, SIREN performs fairly well and outperforms all other baselines, including WIRE. This has also been pointed out by WIRE~\cite{saragadam2023wire}, which states that SIREN catches up in terms of reconstruction quality when scaling up, making it a strong baseline to beat. Our approach, SCONE, not only converges the fastest but also achieves the highest final PSNR.

\begin{figure*}[th]
    \centering
    \includegraphics[width=2\columnwidth]{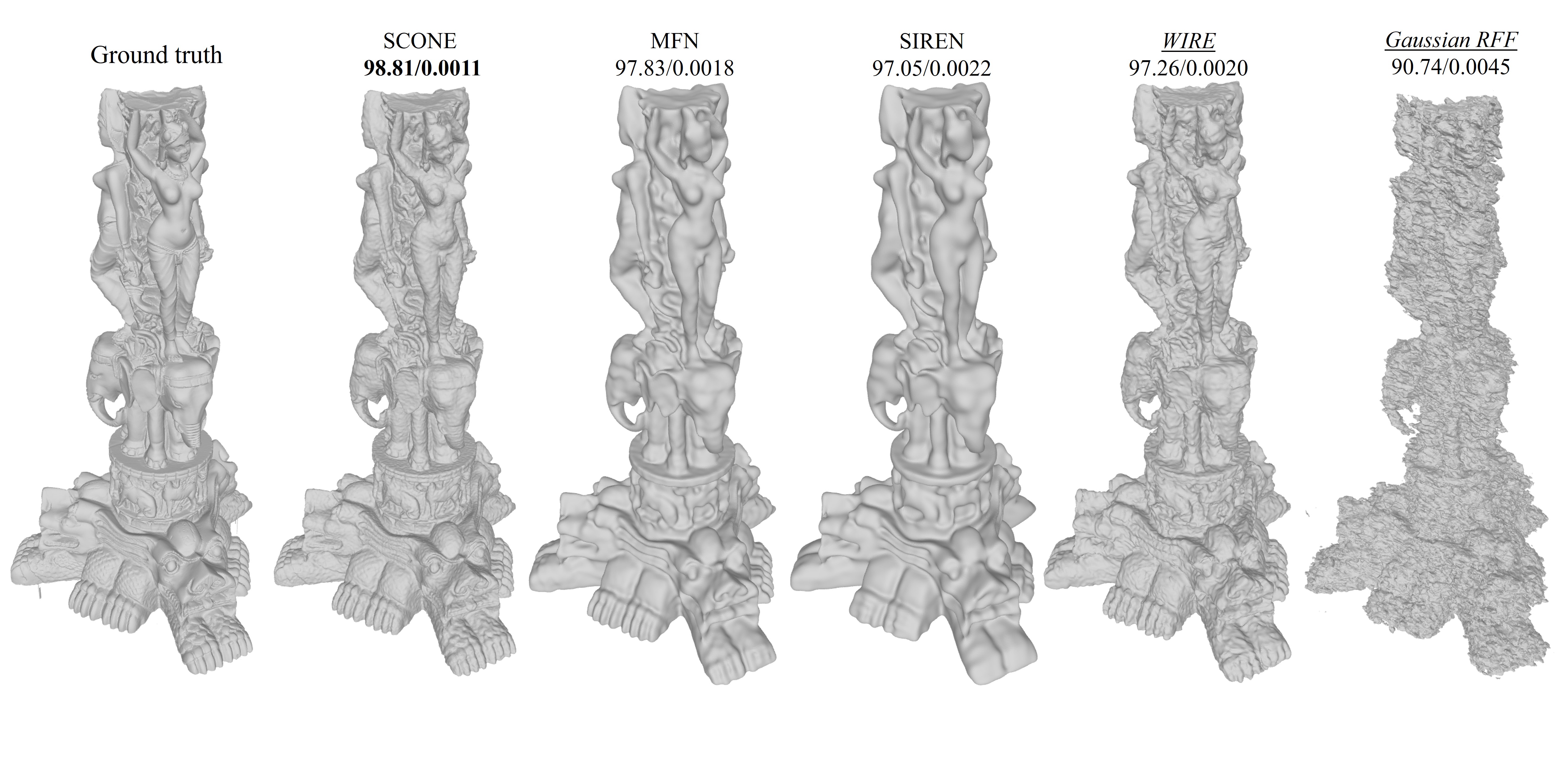}
    \caption{3D shape reconstruction results. All the methods listed are adjusted to have similar model sizes of 400K parameters. All models are trained for 10K steps, except for Gaussian RFF and WIRE cannot converge until 20K steps. SCONE captures more details than other baseline methods, benefiting from collaged Fourier features' design. Best viewed with zoom-in.}
    \label{fig:3d_results}
\end{figure*}

\subsection{Video Representation}
\label{subsec:exp_video}
We further extend our approach from image fitting to video representation tasks. In this case, the input coordinates are now in $[-1, 1]^3$, including an additional temporal domain. All models have the same number of layers and hidden units as in the image-fitting task described above. Following \cite{2020siren}, the video\footnotemark[1] is first downsampled to $512\times512$ with 300 RGB frames. In each iteration, a batch of 262144 random data points is sampled. The mean and standard deviation of PSNR and SSIM across all frames are reported. Table~\ref{tab:video_results} clearly demonstrates that SCONE can be extended to more complex signals, such as video, where data lies in the spatial-temporal domain, achieving the most accurate reconstruction in terms of PSNR and SSIM.

\begin{table}[ht]
\begin{center}
\resizebox{\linewidth}{!}{
\begin{tabular}{lccc}
\toprule
Method & \#Params (K) & PSNR(dB) & SSIM\\
\midrule
Gaussian RFF & 198.1 & $21.74\pm0.71$& $0.35\pm0.02$ \\
SIREN & 199.1 & $26.92\pm0.87$& $0.70\pm0.02$ \\
MFN & 206.3 & $24.02\pm0.73$ & $0.49\pm0.02$ \\
WIRE & 196.1 & $26.90\pm0.78$& $0.67\pm0.02$ \\
\midrule
\textbf{SCONE} (ours) & \textbf{194.0} & \textbf{27.41 $\pm$ 0.83} & \textbf{0.70 $\pm$ 0.01}  \\
\bottomrule
\end{tabular}
}
\end{center}
\caption{Results on video representation task. The mean $\pm$ standard deviation across all frames is reported. SCONE achieves the best performance in terms of PSNR. It ties with SIREN in SSIM, but with smaller variance across frames.}
\label{tab:video_results}
\end{table}

\footnotetext[1]{https://www.pexels.com/video/the-full-facial-features-of-a-pet-cat-3040808}

\subsection{3D Shape Representation}
\label{subsec:3dshape}
For the 3D shape representation task, we use the Stanford 3D scan dataset \footnote[2]{http://graphics.stanford.edu/data/3Dscanrep/}. The dataset contains point clouds with normals of different 3D objects. Following~\cite{lindell2022bacon}, for each step during training, we extract a batch of points from the original point cloud and add a random noise, which consists of two different levels of perturbation, following Laplace distribution with scales of 1e-1 and 1e-3, respectively. The normals for these perturbed points are approximated by taking the average of normals of the nearest three points in the original point cloud. After that, we use the approximated normals to calculate the signed distance to the surface as SDF (signed distance function) labels. MSE loss is used as supervision criteria. To ensure a fair comparison, we adjust SCONE and all the other baseline methods to have similar model sizes of around 400K parameters. Specifically, we use 8 layers for SCONE and empirically initialize the RFF with $\omega$ being $[70, 70, 60, 50, 40, 40, 30, 30]$. We train all the methods with the learning rate following their original implementations, and for 10K steps. Except for Gaussian RFF~\cite{tancik2020fourier} and WIRE~\cite{saragadam2023wire}, which are slower to converge, we train them for 20K steps. For evaluation, we uniformly sample $512\times 512 \times 512$ points in the cube $[-0.5, 0.5]^3$. The networks then predict the signed distance for these points, outputting a tensor of shape $512\times 512 \times 512$, each element denotes the signed distance to the 3D shape plane. 

IoU and Chamfer Distance (CD) metrics are leveraged to quantify the reconstruction results. To compute the IoU scores, we transform the output tensor into an occupancy grid by setting the voxel value to 0 if the corresponding SDF value $>0$ (i.e., outside the 3D shape), and 1 if SDF $ \leqslant 0$ (i.e., inside the 3D shape). IoU is then computed between the ground truth and the predicted occupancy grids. For CD, we use the Trimesh library~\cite{trimesh} to convert the SDF tensor into a 3D mesh and take all the vertices from a mesh as a point cloud for calculating CD. 

Qualitative results are shown in Fig.~\ref{fig:3d_results}. As can be observed, our method captures details much better than the other baseline methods, yielding a 98.81\% IoU and 0.0011 CD, 0.98\% and 63\% higher than the second best. This observation validates our motivation to use collaged Fourier features to improve the model's expressiveness and ability to represent local details. Nonetheless, it is worth stressing the baselines' results are slightly different from the values reported in previous studies because we aligned all the methods under the same experiment settings of (1) model sizes of 400K and training step of 10K for fairness, (2) random sampling points for training, (3) IoU and CD calculating method. The only two exceptions are WIRE~\cite{saragadam2023wire}, and Gaussian RFF~\cite{tancik2020fourier}, who cannot converge and learn any shape within 10K steps. Therefore, for these two methods, we train them for a long period of 20K steps. However, even being trained with double iterations, the IoU and CD results are still relatively low compared to other methods.

\begin{figure}[!ht]
    \centering
    \includegraphics[width=0.95\columnwidth]{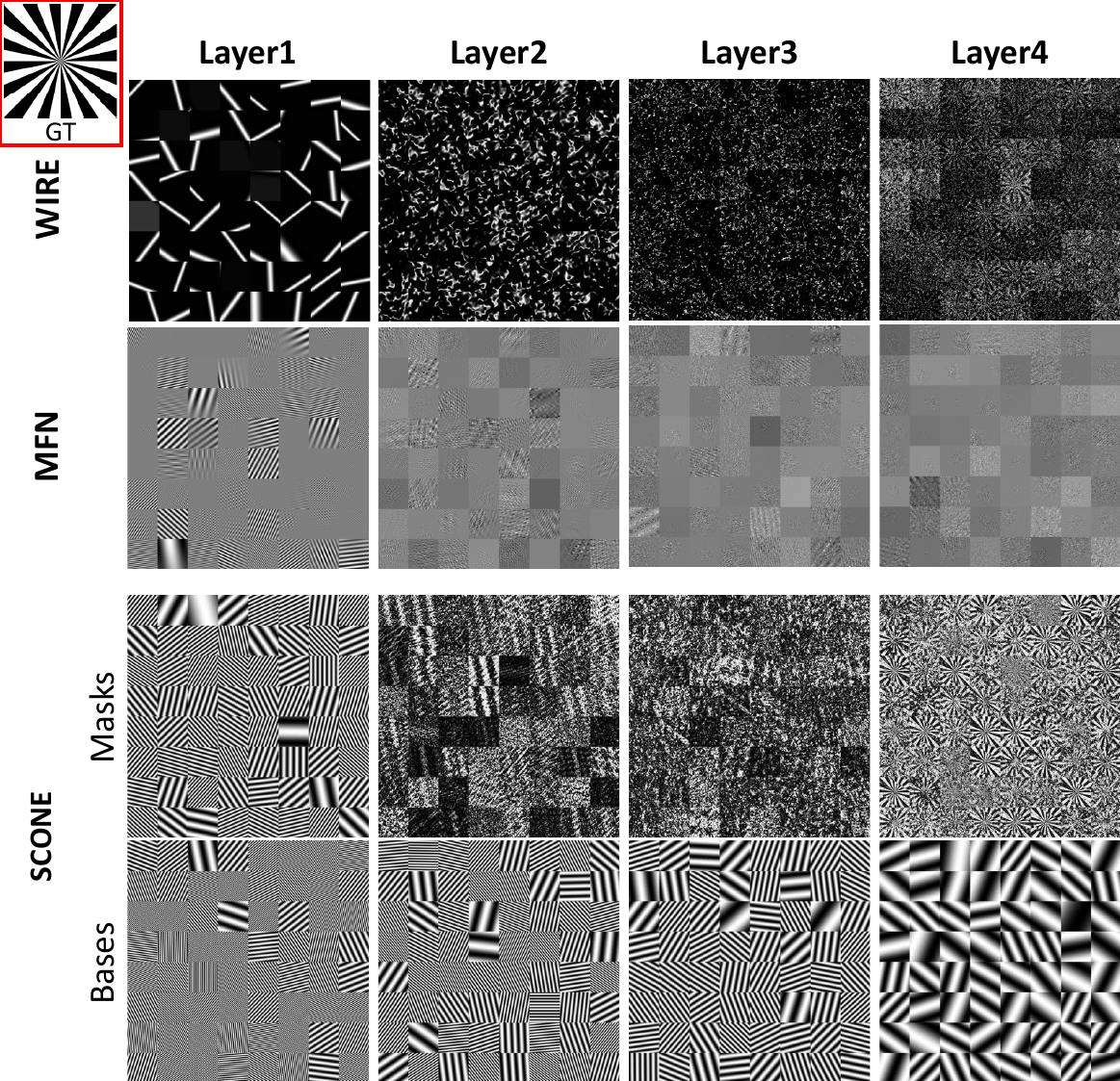}  
    \caption{Visualization of activations from hidden layers. The Siemens star test image is used to highlight the mechanisms that differentiate SCONE from other INRs with wavelet filters or nonlinearities. Each layer consists of 64 units arranged in an $8\times8$ grid, with output of each unit having the same size as the input image ($256\times256$). For WIRE, only the real parts of the activation outputs are shown.}
    \label{fig:activations}
\end{figure}

\subsection{SCONE Layer Visualization}
\label{subsec:mask_visual}
Following WIRE~\cite{saragadam2023wire}, we visualize the layer activations of INRs being trained on the Siemens star test image for 10k iterations, which contain all spatial frequencies and orientations. We compare SCONE with WIRE and GaborNet MFN to highlight their differences in achieving space-frequency locality. Here, each model has 4 hidden layers with a dimension size of 64. The PSNR of SCONE, WIRE, and MFN are 38.47dB, 34.10dB, and 28.26dB, respectively. Fig.~\ref{fig:activations} visualizes the outputs of each hidden layer of the respective models. It is apparent that WIRE and MFN, both utilizing Gabor wavelets, produce sparse activations that are compact in space. In contrast, each base of SCONE is global in nature, spatially spanning the entire channel. The spatial mask then dynamically assigns Fourier bases to the appropriate spatial locations based on the geometric structure of the image. This leads to potentially higher utilization of the channels.

\begin{figure}[ht]
    \centering
    \includegraphics[width=1\columnwidth]{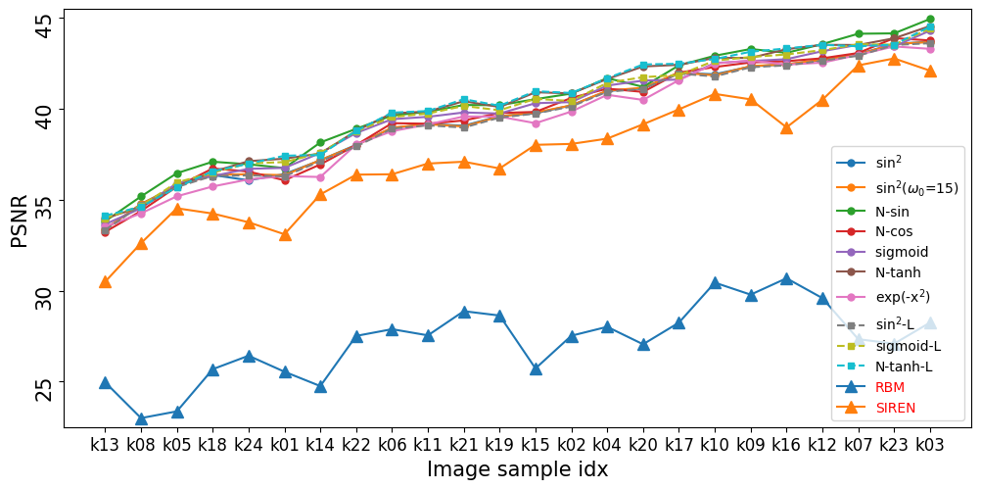}
    \caption{Ablation studies for activation function $\sigma$ on the Kodak dataset. Image samples are sorted by difficulty (average PSNR) for better visualization. All variants of SCONE yield similarly high PSNR and outperform the baselines.}
    \label{fig:act_psnr}
\end{figure}

\section{Ablation Study}
\label{subsec:ablation}

In this section, we conduct ablation studies with the image fitting task using the Kodak dataset to address two questions: (1) Does the performance gain truly come from spatially collaging Fourier bases? (2) Is SCONE robust to different choices of activation functions?

\subsection{Comparison with Random Binary Mask}
 To justify the effectiveness of our design choice, we compare our method to the baseline that uses randomly generated binary masks instead of learnable soft masks in SCONE. The baseline method can be formulated as:
\vspace{-5pt}
\begin{align*}
         \mathbf{z}_{\ell} &=  \mathcal{M}_{\ell} \circ \mathbf{W_{\ell-1}}\mathbf{z}_{\ell-1} \circ g_{\ell}(\mathbf{x};\omega_{\ell}) 
\end{align*}
\vspace{-14pt}

\noindent where $\circ$ denotes element-wise multiplication, $\mathcal{M_{\ell}}\in \{0, 1\}^{D_{\ell}\times H\times W}$ is the random binary mask (RBM) for the $\ell$-th layer, $D$ is the channel dimension, and $H$ and $W$ are the spatial sizes of the image. RBMs are randomly initialized for each layer and fixed during training.

As shown in Fig.~\ref{fig:act_psnr}, the RBM method has much lower PSNR scores compared to SCONE and SIREN, which justifies that SCONE's improvement in performance indeed originates from collaging Fourier bases with learnable spatial masks.

\subsection{Effect of the Choice of Activation Functions}
\label{subsec:activation_ablation}
In this section, we explore the effects of different activation functions on SCONE for generating spatial masks. As mentioned above, a soft mask with values ranging from 0 to 1 is expected. Several activation functions, such as $\textit{sigmoid}$, $\exp(-x^2)$ and $\sin^2$ satisfies this property. For other activation functions such as $\sin$, $\cos$, $\tanh$, whose outputs are in finite range, we normalize the output to $[0, 1]$ by min-max normalization (e.g., $\frac{1}{2}\tanh(\cdot)+0.5$).

Fig.~\ref{fig:act_psnr} compares the performances of different activation function choices on the Kodak dataset. The prefix `N-' denotes min-max normalization, and the `-L' postfix indicates that the input is multiplied by a trainable scaler before applying the activation function, allowing for an adjustable input scale. The $x$-axis in Fig.~\ref{fig:act_psnr} represents the data samples sorted by their average PSNR score over all methods, from easy to difficult, for better visualization. 

It is evident in Fig.~\ref{fig:act_psnr} that SCONE is not limited to specific activation functions, but is robust to different activation choices. SCONE with all activation choices yields significantly higher PSNR scores than the baseline SIREN. Additionally, the differences between various activation choices of SCONE are also small. In contrast, SIREN and RBM exhibit a lot of fluctuation from sample to sample, while the SCONE family shows a more consistent performance across simple and difficult images. Furthermore, the learnable activations (indicated by the `-L' postfix) can bring small improvements to some activation functions (e.g.,  $\textit{sigmoid}$, $\tanh$), although these improvements are not substantial.

In the above experiment section,  we select $\sin^2$ due to its favorable features: (i) output range being $[0,1]$ acting as a soft mask; (ii) periodic function with smooth and continuous gradient, which is intuitively connected with Fourier features. However, it is important to note that our SCONE is not limited to $\sin^2$ and is highly robust to different choices of activation functions. In fact, for SCONE, all activation functions, including $\sin^2$, yield similarly good results and significantly outperform the baseline methods. These results validate our motivation for collaging Fourier bases and demonstrate that SCONE exhibits good generality and robustness across different activation choices.

\section{Conclusion}
\label{sec:conclusion}
This paper presents Spatially Collaged Coordinate Networks (SCONE) as a novel solution to overcome the shortcomings of existing INRs whose universal basis functions can lead to subpar performance and undesirable artifacts. In contrast, SCONE exploits learnable spatial masks to assign particular Fourier bases to their respective regions, thus enabling a more accurate representation of complicated signals. In-depth analyses of SCONE's performance on various INR tasks, such as image fitting, video representation and 3D shape representation, demonstrate excellent reconstruction quality. Specifically, SCONE achieves over 3dB PSNR improvement in image fitting and establishes new SOTA performance in 3D shape representation for general INRs with 99\% IoU and 0.0011 Chamfer Distance, surpassing all competing methods by a large margin.

\section{Acknowledgement}
This work was supported in part by the Theme-based Research Scheme (TRS) project T45-701/22-R and in part by the General Research Fund (GRF) project 17209721 of the Research Grants Council (RGC), Hong Kong SAR.

\clearpage
\bibliography{ref}
\bibliographystyle{aaai24}

\clearpage
\onecolumn
\section{Appendix}
This section includes more detailed experiment results of (1) Quantitative results on the Kodak dataset, (2) Qualitative results for image fitting, and (3) Qualitative results for 3D shape representation.

\subsection{Quantitative Results for the Kodak Dataset}
Due to the page limits, only part of the results for the image fitting task on the Kodak dataset are presented in the manuscript. In this section, we provide a comprehensive report by showcasing all of the quantitative results of SCONE as well as other baselines. As shown in Tab.\ref{tab:all_kodak_results}, all SCONE results are significantly higher than the baseline methods, regardless of which activation function $\sigma$ is used.

\begin{table*}[!ht]
\centering
\resizebox{1\columnwidth}{!}{%
\begin{tabular}{ccccccccccc}
\toprule
\multirow{2}{*}{Images} & \multicolumn{4}{c}{SCONE} & \multirow{2}{*}{SIREN} & \multirow{2}{*}{Gassusian RFF} & \multirow{2}{*}{MFN} & \multirow{2}{*}{WIRE} & \multirow{2}{*}{RBM}\\
\cmidrule(lr){2-5} 
& \multicolumn{1}{c}{$\sin^2$} & \multicolumn{1}{c}{$\textit{sigmoid}$} & \multicolumn{1}{c}{N-$\tanh$}
&  \multicolumn{1}{c}{N-$\tanh$-L} & & & & \\
\midrule
k01 & 36.41/0.96 & 36.75/0.96 & 37.26/0.97 & 37.48/0.97 & 33.04/0.94 & 29.92/0.87 & 28.78/0.87 & 37.49/0.98 & 25.48/0.71 \\
k02 & 40.19/0.96 & 40.37/0.96 & 40.85/0.96 & 40.93/0.96 & 38.01/0.94 & 35.38/0.88 & 34.60/0.91 & 39.39/0.95 & 27.46/0.55 \\
k03 & 43.69/0.98 & 44.30/0.98 & 44.55/0.98 & 44.58/0.98 & 42.04/0.97 & 38.07/0.94 & 37.49/0.95 & 39.19/0.94 & 28.18/0.58 \\
k04 & 40.98/0.97 & 41.28/0.97 & 41.64/0.97 & 41.74/0.97 & 38.30/0.95 & 35.46/0.90 & 35.51/0.93 & 38.23/0.95 & 27.95/0.62 \\
k05 & 35.80/0.96 & 35.91/0.96 & 35.73/0.96 & 35.74/0.96 & 34.47/0.96 & 29.06/0.87 & 28.63/0.91 & 31.76/0.93 & 23.30/0.65 \\
k06 & 38.94/0.97 & 39.43/0.97 & 39.75/0.97 & 39.85/0.97 & 36.34/0.95 & 31.94/0.89 & 31.39/0.91 & 37.67/0.97 & 27.82/0.69 \\
k07 & 43.01/0.98 & 43.52/0.98 & 43.51/0.98 & 43.50/0.98 & 42.33/0.98 & 37.29/0.95 & 36.32/0.96 & 37.82/0.95 &27.27/0.64 \\
k08 & 34.76/0.96 & 34.49/0.95 & 34.61/0.95 & 34.68/0.95 & 32.55/0.94 & 26.17/0.80 & 25.73/0.86 & 31.98/0.95 & 22.93/0.67 \\
k09 & 42.33/0.97 & 42.63/0.97 & 43.20/0.97 & 43.20/0.97 & 40.46/0.95 & 36.32/0.91 & 35.58/0.93 & 40.12/0.97 & 29.72/0.69 \\
k10 & 41.90/0.97 & 42.52/0.97 & 42.80/0.97 & 42.80/0.97 & 40.76/0.96 & 36.37/0.92 & 35.38/0.94 & 38.20/0.94 & 30.38/0.73 \\
k11 & 39.17/0.96 & 39.56/0.96 & 39.85/0.97 & 39.94/0.97 & 36.93/0.95 & 31.70/0.91 & 33.16/0.87 & 37.83/0.96 & 27.48/0.65 \\
k12 & 42.68/0.97 & 43.17/0.98 & 43.52/0.98 & 43.58/0.98 & 40.44/0.96 & 36.58/0.91 & 35.94/0.94 & 39.36/0.95 & 29.53/0.66 \\
k13 & 33.38/0.95 & 33.65/0.95 & 33.97/0.95 & 34.18/0.96 & 30.43/0.91 & 26.57/0.79 & 26.07/0.84 & 34.73/0.98 & 24.90/0.73 \\
k14 & 37.17/0.96 & 37.57/0.96 & 37.54/0.96 & 37.52/0.96 & 35.23/0.94 & 31.74/0.89 & 31.49/0.91 & 36.81/0.96 & 24.69/0.61 \\
k15 & 39.76/0.96 & 40.31/0.96 & 40.91/0.96 & 41.03/0.96 & 37.96/0.94 & 34.67/0.89 & 33.06/0.91 & 37.85/0.94 & 25.66/0.47 \\
k16 & 42.45/0.98 & 42.73/0.98 & 43.28/0.98 & 43.38/0.98 & 38.92/0.96 & 35.52/0.92 & 35.13/0.94 & 41.02/0.97 & 30.62/0.75 \\
k17 & 41.98/0.98 & 41.61/0.97 & 42.42/0.98 & 42.54/0.98 & 39.91/0.97 & 35.65/0.92 & 34.24/0.94 & 40.31/0.97 & 28.18/0.64 \\
k18 & 36.36/0.95 & 36.30/0.95 & 36.54/0.95 & 36.63/0.95 & 34.18/0.93 & 30.95/0.86 & 29.69/0.90 & 34.65/0.93 & 25.62/0.62 \\
k19 & 39.60/0.96 & 39.74/0.96 & 40.10/0.97 & 40.23/0.97 & 36.66/0.94 & 31.35/0.90 & 32.97/0.88 & 34.48/0.91 & 28.57/0.70 \\
k20 & 41.12/0.97 & 41.55/0.97 & 42.33/0.97 & 42.50/0.97 & 39.09/0.96 & 35.50/0.92 & 33.78/0.93 & 40.26/0.97 & 26.99/0.57 \\
k21 & 39.05/0.96 & 39.80/0.96 & 40.42/0.97 & 40.59/0.97 & 37.03/0.95 & 33.28/0.90 & 31.87/0.93 & 37.49/0.94 & 28.81/0.70 \\
k22 & 38.02/0.95 & 38.67/0.96 & 38.79/0.96 & 38.85/0.96 & 36.32/0.94 & 32.95/0.88 & 32.12/0.90 & 36.16/0.94 & 27.45/0.65 \\
k23 & 43.59/0.98 & 43.43/0.97 & 43.88/0.98 & 43.59/0.98 & 42.71/0.97 & 38.16/0.93 & 38.81/0.96 & 30.88/0.69 & 27.02/0.50 \\
k24 & 36.07/0.95 & 36.70/0.95 & 37.12/0.96 & 37.04/0.96 & 33.70/0.94 & 30.32/0.84 & 28.73/0.87 & 35.48/0.96 & 26.36/0.67 \\
\midrule
Mean & 39.52/0.96 & 39.83/0.97 & \textit{40.19/0.97} & \textbf{40.25/0.97} & 37.41/0.95 & 33.37/0.89 & 32.77/0.91 & 37.05/0.94 & 27.18/0.64\\
\bottomrule
\end{tabular}%
}
\caption{Quantitative results on all images of Kodak dataset. Scores are reported in the form of PSNR/SSIM.}
\label{tab:all_kodak_results}
\end{table*}

\subsection{Additional Qualitative Results for Image Fitting}
\begin{figure*}[!ht]
    \centering
    \includegraphics[width=.6\columnwidth]{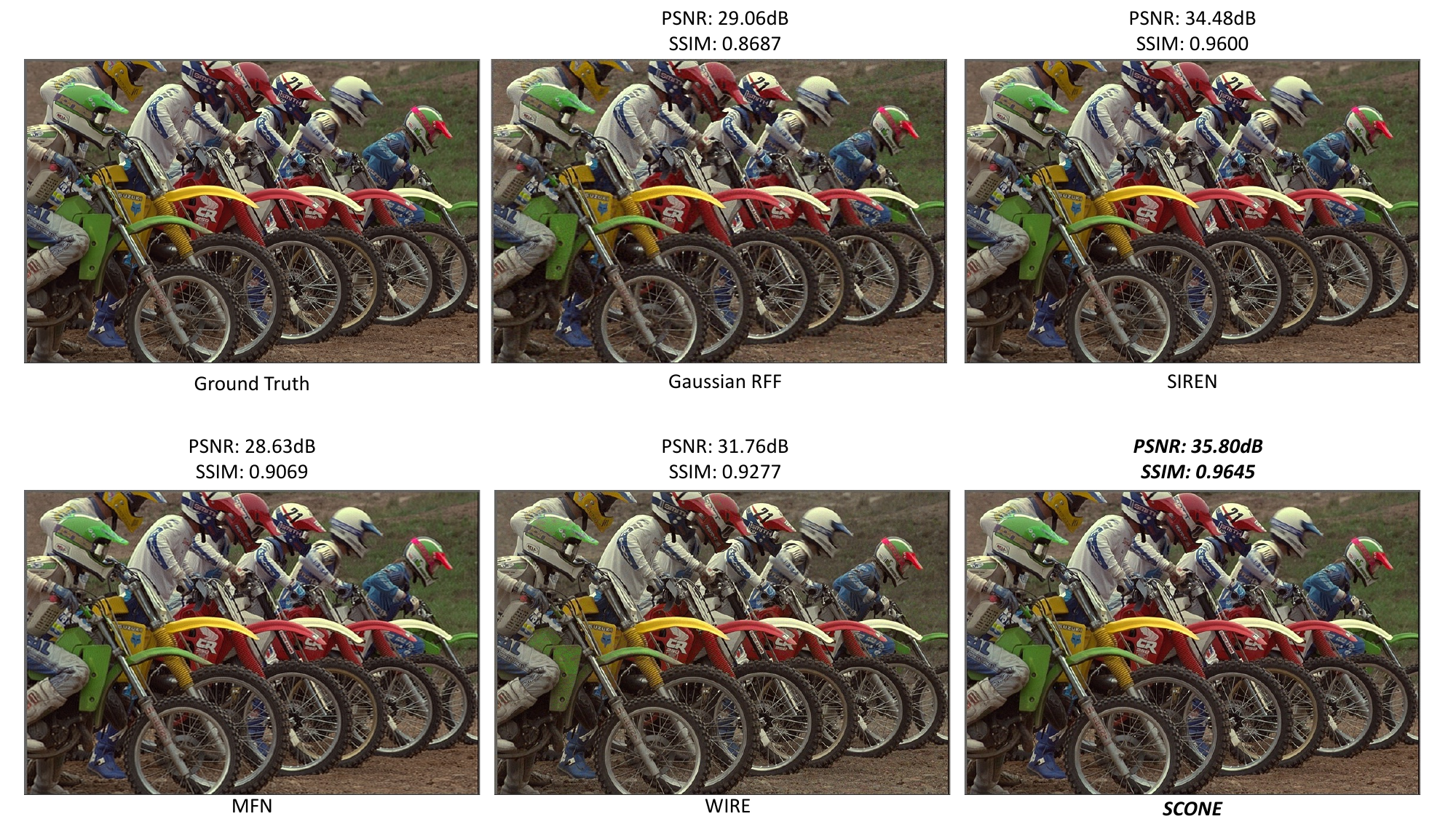}
    \caption{Image fitting result on Kodak05.}
    \label{fig:kodak05_result}

    \centering
    \includegraphics[width=.6\columnwidth]{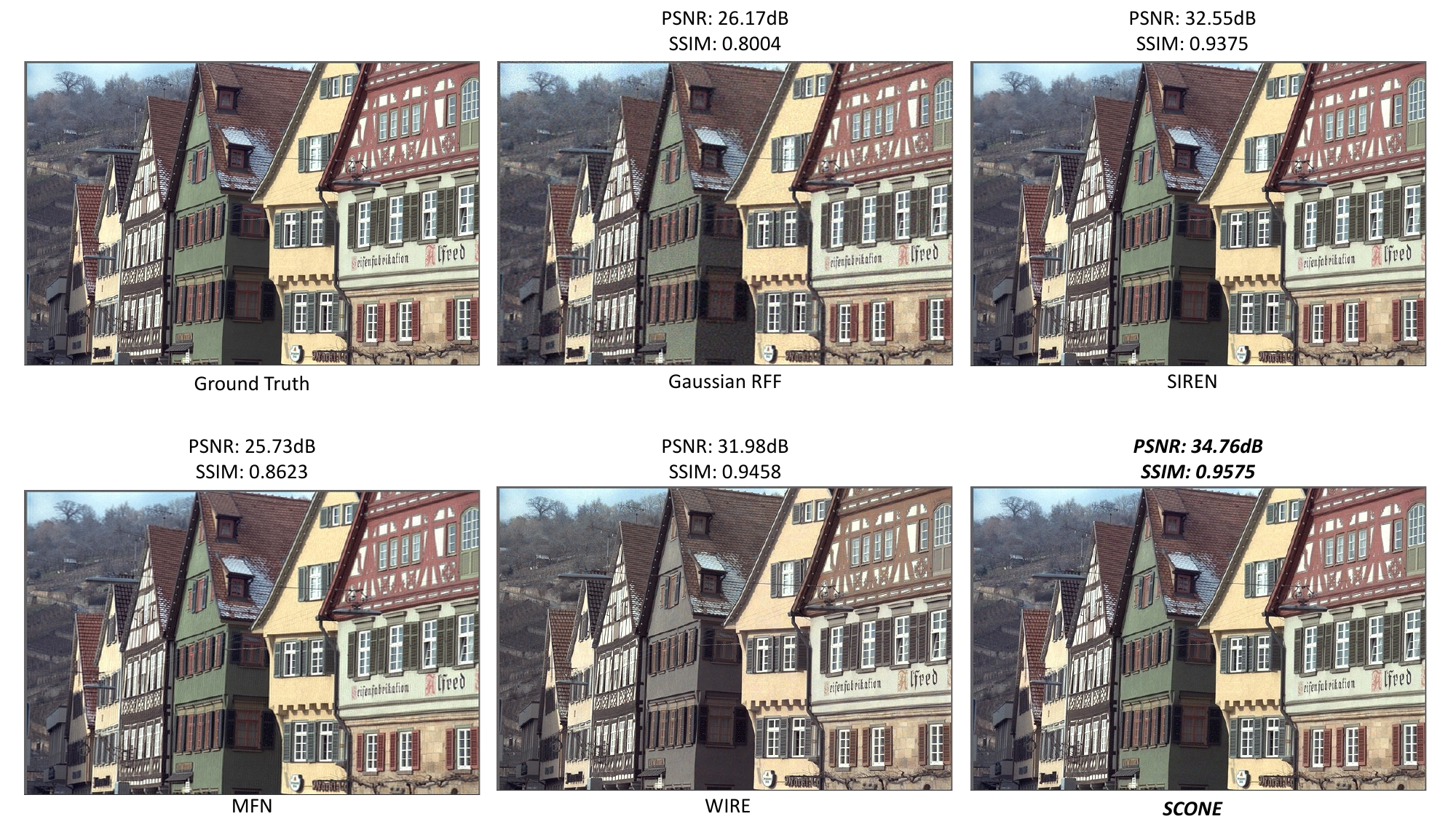}
    \caption{Image fitting result on Kodak08.}
    \label{fig:kodak08_result}

    \centering
    \includegraphics[width=.6\columnwidth]{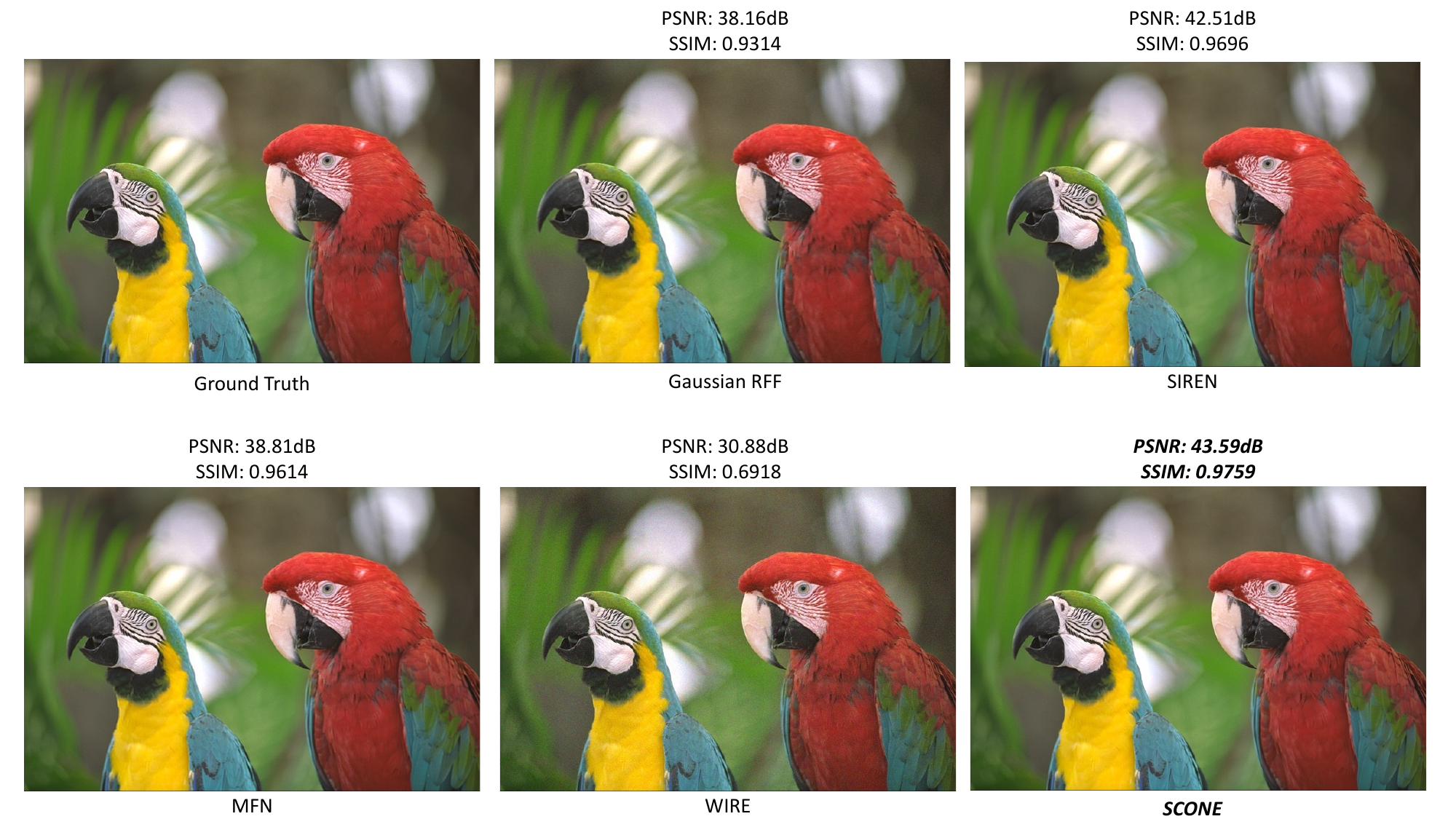}
    \caption{Image fitting result on Kodak23.}
    \label{fig:kodak23_result}
\end{figure*}

Here, we present additional results on image fitting tasks using selected images from Kodak datasets, including Kodak05, Kodak08, and Kodak23. We follow the same settings and hyperparameters as in the main manuscript. Specifically, we use a learning rate of 1e-4 for SIREN and SCONE, 1e-3 for ReLU MLP with Gaussian RFF and WIRE, and 1e-2 for GaborNet MFN. Each model is trained for 10k iterations. For SCONE, we initialize $\omega_0 = 30$ and $\omega_{\ell} = [90, 60, 30, 10]$ for RFFs in respective layers. For SIREN, we set $\omega_0 = 30$. For MFN, we set the weight scale to 1, the input scale to 256, and $\alpha=6$ and $\beta=1$. For WIRE, we set $\omega_0 = 20$ and $s_0 = 20$. For Gaussian RFF, we set $\sigma=10$ and the encoding size to 128. As shown in Fig.~\ref{fig:kodak05_result}, Fig.~\ref{fig:kodak08_result}, and Fig.~\ref{fig:kodak23_result}, SCONE consistently outperforms all other baselines in both PSNR and SSIM, demonstrating its efficacy.

\clearpage

\subsection{Additional Qualitative Results for 3D Shape Representation}

\begin{figure*}[ht]
    \centering
    \includegraphics[width=1\columnwidth]{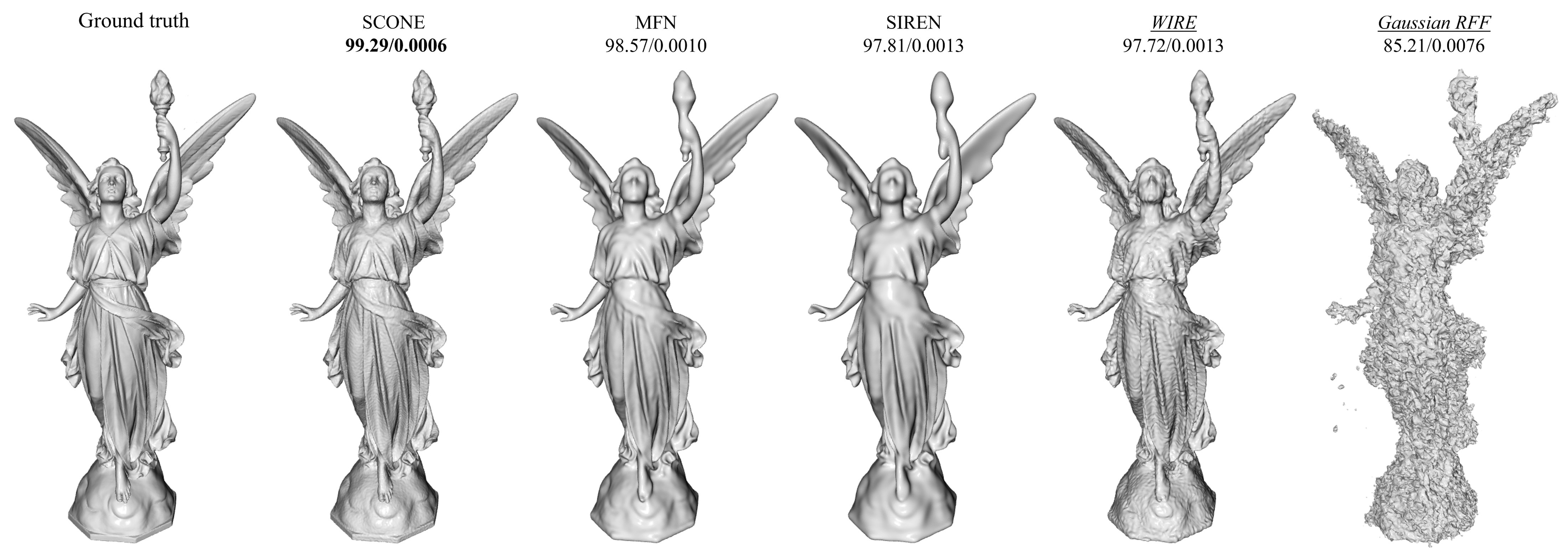}
    \caption{Stanford 3D scan dataset, Lucy result. SCONE yields the best result and is able to capture fine details (e.g., cloth and face). Best viewed with zoom-in.}
    \label{fig:lucy_results}

    \centering
    \includegraphics[width=1\columnwidth]{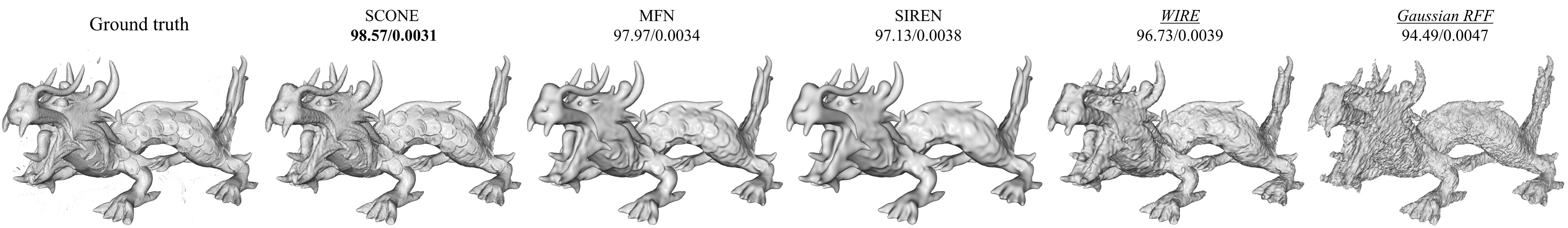}
    \caption{Stanford 3D scan dataset, Dragon result. Despite the presence of some artifacts in the ground truth (e.g., between the horns), SCONE is robust to noise and can generate high-quality results. Additionally, SCONE preserves more details than other methods, such as the eyes and teeth. Please refer to the text discussion for an explanation of the noise in the ground truth. Best viewed with zoom-in.}
    \label{fig:dragon_results}

    \centering
    \includegraphics[width=1\columnwidth]{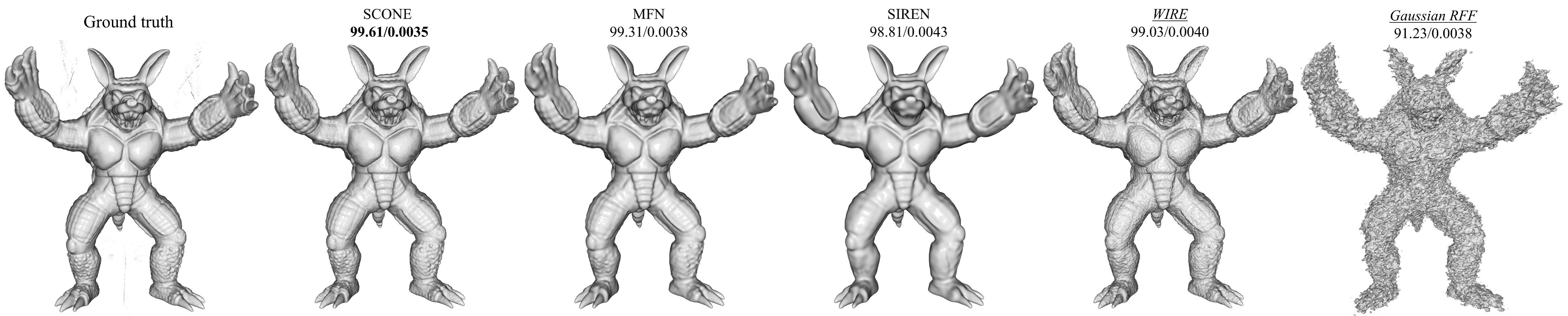}
    \caption{Stanford 3D scan dataset, Armadillo result. Same as in Fig.~\ref{fig:dragon_results}, SCONE preserves better details than other baselines (e.g., teeth) while being robust to the noises shown in the ground truth. Best viewed with zoom-in.}
    \label{fig:armadillo_results}
\end{figure*}

This section presents the remaining experimental results on the Stanford 3D scan dataset, using the same experimental settings as in the main manuscript. As shown in Fig.~\ref{fig:lucy_results}, Fig.~\ref{fig:dragon_results} and Fig.~\ref{fig:armadillo_results}, SCONE significantly outperforms other baselines and preserves finer shape details. All methods are trained for 10K steps, except for Gaussian RFF~\cite{tancik2020fourier} and WIRE~\cite{saragadam2023wire}, which require 20K steps to converge. However, even when trained for double the iterations, these two methods still yield lower accuracy than SCONE as well as some other baseline methods.

Unlike the previous results presented in the main manuscript, the ground truths for the 3D shape `dragon'  and `armadillo' contain some artifacts. The reason for this is that, when constructing the ground truth SDF grids, the normal of each grid center is approximated by averaging the normals of its three nearest neighbors. However, for some grids located between two separate parts of the 3D shape (e.g., between the arm and the head of the armadillo in Fig.~\ref{fig:armadillo_results}), the three nearest neighbors may belong to different parts. As a result, averaging these neighbors leads to inaccurate normal approximation and, consequently, artifacts. However, we find surprisingly that although the ground truth contains noise, almost all of the INR approaches are robust to these outliers and can generate clean results. SCONE, in particular, captures the most details of the ground truth and does not contain the noise.

\end{document}